# Tendon-Driven Soft Robotic Gripper with Integrated Ripeness Sensing for Blackberry Harvesting

Alex Qiu[1], Claire Young[2], Anthony L. Gunderman[3], Milad Azizkhani[1], Yue Chen[2], Ai-Ping Hu[4]

*Abstract*— **Growing global demand for food, coupled with continuing labor shortages, motivates the need for automated agricultural harvesting. While some specialty crops (e.g., apples, peaches, blueberries) can be harvested via existing harvesting modalities, fruits such as blackberries and raspberries require delicate handling to mitigate fruit damage that could significantly impact marketability. This motivates the development of soft robotic solutions that enable efficient, delicate harvesting. This paper presents the design, fabrication, and feasibility testing of a tendon-driven soft gripping system focused on blackberries, which are a fragile fruit susceptible to post-harvest damage. The gripper is both low-cost and small form factor, allowing for the integration of a micro-servo for tendon retraction, a near-infrared (NIR) based blackberry ripeness sensor utilizing the reflectance modality for identifying fully ripe blackberries, and an endoscopic camera for visual servoing with a UR-5. The gripper was used to harvest 139 berries with manual positioning in two separate field tests. Field testing found an average retention force of 2.06 N and 6.08 N for ripe and unripe blackberries, respectively. Sensor tests identified an average reflectance of 16.78 and 21.70 for ripe and unripe blackberries, respectively, indicating a clear distinction between the two ripeness levels. Finally, the soft robotic gripper was integrated onto a UR5 robot arm and successfully harvested fifteen artificial blackberries in a lab setting using visual servoing.**

## I. INTRODUCTION

IN 2005 an estimated 38% of the world's land area was occupied by agricultural needs alone [1]. With the global population projected to reach 8.5 billion by 2030 [2], it is estimated that without significant technological development in the agricultural sector, 50% of all land will need to be devoted to agriculture to maintain food consumption levels. Furthermore, by 2050, an estimated 70% increase in food production will be needed to support a projected 9.1 billion world population [3]. To meet these growing needs, substantial improvements in agricultural technology have been made, specifically the development and applications of machinery and automation. These advances have allowed for lowered production costs, reduced time spent on strenuous labor, increased fresh produce quality, etc. [4]. Among the various activities involved in the agricultural cycle, crop harvesting is both a time-consuming and labor-intensive task that constitutes a relatively large portion of farm expenditure. For example, in the blackberry industry, harvesting hours consist of 40-50% of the total man hours in maintaining and harvesting a crop [5]. As a result, in one study, the cost of harvesting contributed to 56% of the total cost of bringing a blackberry to market [6]. These considerations have pushed automated harvesting to the forefront of extensive research within agricultural robotics [7].

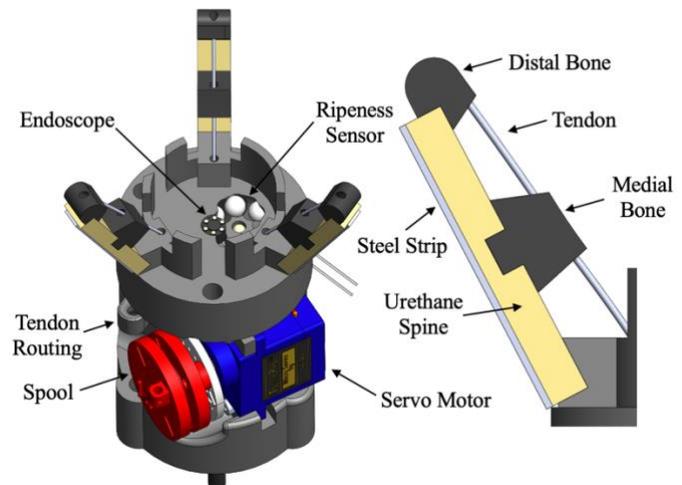

Fig. 1. Developed soft robotic gripper assembly with integrated ripeness sensor and eye-in-hand endoscope camera.

Numerous methods for automated harvesting have been developed for various crop types, such as apples [8-10], tomatoes [11-13], strawberries [14-16], onions [17], etc. that have been successfully implemented. For example, Brown et al. [18] developed a mechanized blueberry harvester that was capable of harvesting 68% of blueberries with no post-harvest damage, compared to an average of 77% for traditional hand-harvesting. Xiong et al. [14] developed a robotic strawberry harvester that was able to position itself with respect to the berry in five seconds. These methods utilize rigid robotics and mechanization with shaking and cutting motions that can be extremely effective for harvesting robust crops in great quantity. However, when it comes to soft, fragile fruits, such as

This research was funded the Georgia Tech IRIM Seed Grant. Corresponding author: Alex Qiu.

[1]A. Qiu and M. Azizkhani are with the School of Mechanical Engineering, Georgia Institute of Technology, Atlanta, 30338, USA (e-mail: aqiu34@gatech.edu, mazizkhani3@gatech.edu)

[2]C. Young, and Y. Chen are with the Biomedical Engineering Department, Georgia Institute of Technology/Emory, Atlanta, 30338, USA (e-mail: cyoung301@gatech.edu, yue.chen@bme.gatech.edu).

[3]A. Gunderman is with the School of Electronics and Computer Engineering, Georgia Institute of Technology, Atlanta, 30338, USA (e-mail: agunderman3@gatech.edu)

[4]A.-P. Hu is with the Intelligent Sustainable Technologies Division, Georgia Tech Research Institute, 640 Strong St., Atlanta, 30318 USA (e-mail: ai-ping.hu@gtri.gatech.edu).

blackberries and raspberries, these methods are ineffective, as post-harvest damage severely impacts their marketability, resulting in loss [19]. As a consequence, these fruit are commonly harvested by hand to prevent damage, resulting in significant labor costs and berry losses due to labor skill [20].

This motivates the implementation of soft robotics solutions, such as soft robotic grippers, which can enable gentle, automated harvesting while mitigating post-harvest damage [21]. Many studies have been conducted for soft robotic grippers with various crop types, ranging from eggplants, cucumbers, and broccoli to apples, oranges, bananas, etc., which all vary case by case in gripper designs due to a variety of factors [22]. For example, Hayashi et al. [23] developed a soft robotic gripper system that could both softly grasp eggplants and cut the peduncles off with a 62.5% harvesting success rate. Morales et al. [24] created a suction cup based soft robotic gripper targeting vegetables such as onions and artichokes, which was able to harvest and place the crops at 60 cycles per minute. Another study done by Ariyanto et al. [25] looked into the development of a pneumatic powered three-finger soft robotic gripper capable of grasping grapes without destruction. However, of these crop types, blackberries are arguably the most difficult to harvest, precluding applicability of prior gripper designs. This is due to: (i) blackberries are a delicate fruit consisting of fragile drupelets that encompass a soft tissue receptacle [26], (ii) blackberry canopies have abundant leaves and stems that can occlude berries easily at multiple different angles, and (iii) blackberries occur in clusters, causing difficulty in isolating one blackberry from another during harvesting.

To our knowledge, there has been only one study on robotic blackberry harvesting. This was our most recent work, where we developed a tendon-driven soft robotic gripper with force-feedback capable of harvesting blackberries with a 95.24% detachment success rate and 16% red drupelet reversion rate (post-harvest disorder affecting blackberry marketability [19, 27]) in testing [28, 29]. Although this gripper was successful in detaching blackberries, it was limited in two respects: (i) it had a large form factor, preventing blackberry isolation in clusters and colliding with the blackberry canopy and (ii) the gripper was not implemented on a robotic system (i.e., a robotic arm).

In addition to the listed limitations, our prior gripper (along with the majority of those already cited) had no method for ripeness detection or alternative ripeness detection methods to RGB computer vision-based perception. Alternative ripeness detection methods are especially important for fruits like blackberries because of how similar near-ripe and ripe blackberries are in color. The need for these methods have been justified by several studies, such as Beghi et al. [30], who showed that near-ripe and ripe blueberries have virtually indistinguishable visible color appearances.

Only a few studies have been conducted for integrated ripeness sensing in soft robotic grippers that rely on more than just RGB computer vision-based perception. Hanson et al. conducted a study developing reflectance-based ripeness sensing integrated into a soft robotic gripper with a processing pipeline for spectral readings [31]. Another study for integrated ripeness sensing into soft robotic grippers was done by Cortés et al. [32] where the quality of mangoes was detected using reflectance readings. While these studies were successful in showing the ripeness sensing capabilities, they lacked gripper harvesting feasibility studies. This project thus aims to address both gripper harvesting capabilities and ripeness sensing, combining what current studies in soft robotic automated harvesting lack.

In this paper, we present a soft robotic gripper with integrated blackberry ripeness sensing and visual servoing capabilities for the harvesting of blackberries, presenting for the first time spectral ripeness values and plant retention forces for ripe and unripe blackberries, both of which can be used for ripeness detection. Two manual field tests were performed using the gripper system and a full feasibility test on a UR5 was performed in the lab, analyzing arm manipulation via visual-servoing and blackberry harvesting. Our goal is to create a robust, autonomous harvesting system free from human intervention for blackberries, with the hopes of paving the way for implementation of soft robotic systems into the agricultural harvesting process. The paper is organized as follows. Section II lays out the design objectives for this project and describes the gripper base design, finger design and fabrication, ripeness sensor, and robotic arm integration. Section III describes the experimental methods for reflectance detection, field tests, and UR5 arm integration test. Section IV discusses the results. The paper is concluded in Section V.

## II. METHODS AND MATERIALS

### A. Design Objectives

To effectively meet the goals of autonomous harvesting of blackberries, several key design objectives were identified.
1. The gripper must be a soft robotic gripper that conforms to the shape being grasped, providing a distributed load on the blackberry surface.
2. The gripper must allow the use of image-based visual servoing (IBVS) [33], which requires an eye-in-hand camera for image feedback. IBVS will allow for localization of blackberries with respect to the gripper and enable closed loop control for approaching them.
3. The gripper must allow for integration of a near-infrared (NIR) based ripeness sensor. The ripeness sensor will be used for blackberry ripeness detection for harvesting.
4. The gripper must be lightweight and have a small form factor for easy integration with a given robotic arm. This will allow maneuverability in the dense blackberry canopy.
5. The robotic harvesting system must be able to autonomously harvest blackberries with diameters ranging from 17 mm to 31 mm in diameter [34].

### B. Gripper Base Design

The soft robotic gripper base (Fig. 1) was designed to incorporate three different elements: (i) an eye-in-hand camera for visual servoing, (ii) a NIR ripeness sensor for blackberry ripeness detection, and (iii) an actuation mechanism for finger deformation and blackberry grasping. The diameter of the gripper base was selected to incorporate these elements (45 mm). The endoscope camera (DPNKJ0015, DEPSTECH, USA) chosen is a 5.5 mm OD USB camera with focal length of 1 mm and an image resolution of 1280×720 pixels. The endoscope was incorporated into the gripper base via an inner lumen concentric with the outer diameter of the gripper base, enabling a known transformation between the robot base and the robot end-effector for visual servoing. A 9g 180° rotation micro servo (TS90A, TianKongRC, China), in conjunction with a custom 3D printed 14 mm diameter spool and 8-pound fishing line (tendon) were used for finger retraction. A 14 mm spool diameter was chosen to allow for an adequate range of finger retraction, from a relatively wide starting configuration to a tight grasping position.

### C. Finger Design and Fabrication

The finger design is inspired by continuum robots used in medical applications [35-37]. This design deviates from our prior work by enabling the use of a smaller servo motor by replacing the internal underactuated tendon design with an external underactuated tendon design. The finger consists of a single continuous bending section comprised of Vytaflex™ urethane rubber 32 mm long and 7.6 mm wide, with a shore hardness of 20A, enabling shape conformation of the gripper to irregular shapes. Each finger has two "bones." The medial bone separates the bending section into two bending sections and the distal bone acts as a termination point for the tendon. The bones of the finger are made of 3D-printed PLA and the ventral side of the bones were covered with urethane to provide a high friction, soft touchpoint to ensure effective grasping of the blackberry. The finger tendon was routed from the spool through the gripper base and then through the medial bone. The tendon consisted of 8-pound fishing line (BGQS8C-15, Berkley, USA), enabling flexibility and abrasion resistance, which improved fatigue life. All three tendons converge at a single point and join as a combined line to wrap around the servo spool, which allows for constant finger retractions with one actuator. A 4.75 mm long and 0.076 mm thick 301 stainless steel strip (9293K111, McMaster, USA) was added to the spine of each finger, increasing the lateral stiffness to avoid disturbances by the plant foliage as well as gripper retention force.

The fingers were equidistantly spaced around the circumference of the gripper to ensure an even loading distribution on the blackberry [38]. The base of the finger was 22 mm from the center axis of the gripper to ensure the finger body did not interfere with the endoscopic camera or ripeness sensing. The finger length and opening angle were selected through rapid prototyping in SolidWorks to incorporate design requirement (2) and (5). The opening angle of the gripper (60°) was selected based on the chosen finger length to ensure that the finger was outside of the endoscopic camera's FOV (80°) (Fig. 2). The finger length was maximized to enable stable grasping of a 31 mm diameter blackberry, subject to the constraint that the fingertips should not touch when closed around a 17 mm diameter berry. Using the piece-wise constant curvature assumption commonly used in continuum robots [39] and assuming each segment bends with the same curvature subject to a given tendon retraction, we used a neutral finger length of 32 mm.

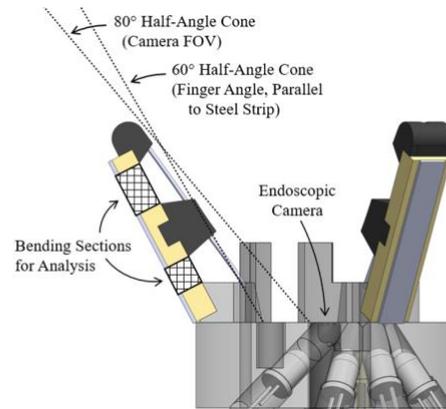

Fig. 2. Finger and camera FOV half angle cone positioned with respect to gripper showing design decisions to keep finger geometry from obstructing camera FOV. Bending sections of finger considered to be undergoing constant curvature bending.

### D. Ripeness Sensor

The NIR ripeness sensor consisted of seven low-cost 870 nm LEDs (LED870E, Thorlabs, USA) and one 850 nm photodiode (365-2015-ND, TTElectronics/Optek Technology, USA), and utilizes reflectance for ripeness detection [30, 40]. 870 nm was chosen as the wavelength because through testing it yielded the highest amount of reflection and differentiation between ripe and unripe blackberries. The photodiode is connected in a voltage divider circuit, with a wire connected to an analog input pin on the Arduino for detecting voltage changes. A resistor value of 560 kOhm was found to be most optimal in allowing for the largest sensitivity range for voltage readings. The photodiode takes reflectance readings before and after the LEDs illuminate the blackberry, calculating the reflectance as the difference of these two values (analog voltage readings from Arduino). The LEDs are inserted at a 35-degree incident angle, which was found to yield the maximum reflectance difference for blackberries using standalone sensor housing units as shown in Fig. 3. This design was chosen due to how the reflectance modality relies on the receptivity of light reflected off the outer surface and insides off the blackberries.

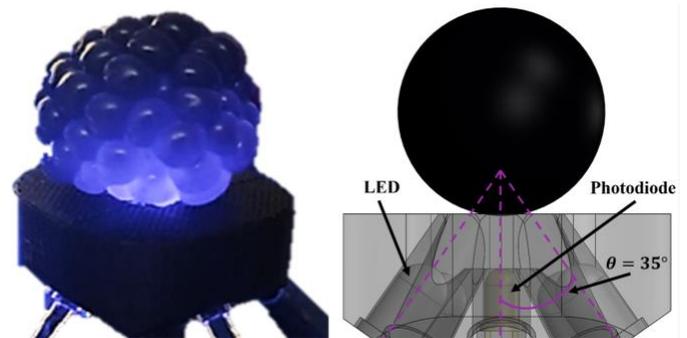

Fig. 3. Standalone ripeness sensor housing unit for finding optimal design parameters for maximized reflectance. Note that with the 870 nm wavelength is not visible light to human eye, but with cameras is detectable as seen in this figure.

### E. Robotic Arm Integration

The soft robotic gripper system was integrated onto a UR5 6-DOF robotic arm (Universal Robots, USA) for a benchtop lab feasibility study with artificial blackberries. UR5 joint commands were provided through Ethernet socket communication via Python code executed on a Windows laptop. Gripper commands were provided through serial communication with an Arduino via the same Python code. OpenCV was used for endoscopic image processing and a pre-trained neural network YOLO v5s was used for real-time artificial blackberry detection to aid with IBVS.

IBVS is a technique used in robotics that utilizes image feedback to learn a mapping between robot arm space and camera space, without the need for camera calibration or a robot arm model [33]. This project applied the eye-in-hand case utilizing the endoscopic camera rigidly attached to the center of the gripper. The visual servoing process can be broken into two main steps: (i) image Jacobian calibration and (ii) arm adjustments. The image Jacobian was calculated by jogging the end-effector known cartesian distances and recording the change in image space coordinates of a pre-chosen object for the x and y axis. In this case, the chosen object was always the left-most detected blackberry in the image, which was detected using YOLO v5s. Once the image Jacobian was known, the arm would first make 2D adjustments to center the blackberry in the image space and then approach the blackberry by 50% of the estimated distance from the blackberry. Depth was estimated by utilizing OpenCV's built-in ArUco tag library depth estimation function. Using the YOLO model to return a bounding box surrounding the target blackberry, an average tag size corresponding to the average of the width and length of the bounding box was calculated. This was then inputted as a parameter into the ArUco tag depth estimation function where a corresponding depth estimation was returned. After each approach, the procedure would repeat until the end-effector converged to the location of the blackberry, where the gripper would be in position for harvesting. This protocol can be seen in Fig. 4.

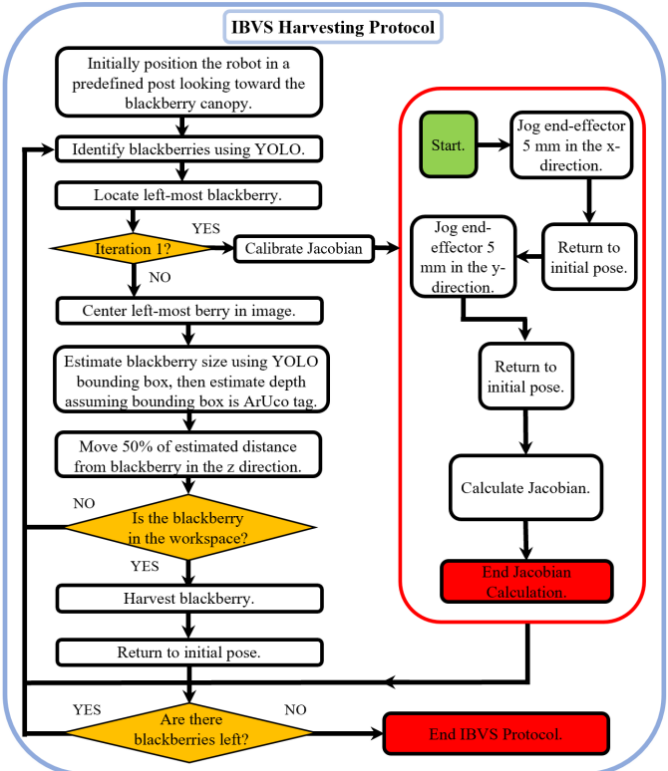

Fig. 4. IBVS protocol for harvesting blackberries using the UR5 robot.

## III. EXPERIMENTAL METHODS

### A. Reflectance Detection

Ripeness sensor calibration required the testing and analysis of reflectance values for fully ripe blackberries (Fig. 5(left)) and unripe blackberries (Fig. 5(right)). Ripe and unripe blackberries differ in drupelet color, firmness, size, and translucence, resulting in differing reflectance values. Reflectance detection is used in our field test to validate blackberry ripeness. Thus, reflectance calibration must be performed first. Calibration consisted of using the integrated NIR ripeness sensor (Fig. 5). The LEDs were powered using a supply voltage of 1.85 V, providing a consistent light intensity. The resulting steady-state analog output voltage in ambient lighting conditions was recorded as the initial reflectance $R_0$. Note that ambient light associated with the time of day was observed to have a negligible impact on reflectance readings. To obtain the ripe $R_{ripe}$ and unripe $R_{unripe}$ reflectance values, a ripe and unripe blackberry were placed individually on the NIR sensor. The gripper was allowed to grasp the blackberry to ensure that the blackberry was centered on the NIR sensor. The resulting steady-state analog output voltage in ambient lighting conditions was recorded as the reflectance as $R_b$. The reflectance for the tested blackberry was then calculated by $R_f = R_b - R_0$. Note that the ripeness sensor uses a pre-determined physical offset of 5 mm from the gripper palm, thus the target blackberry needs to be touching the palm before taking measurements.

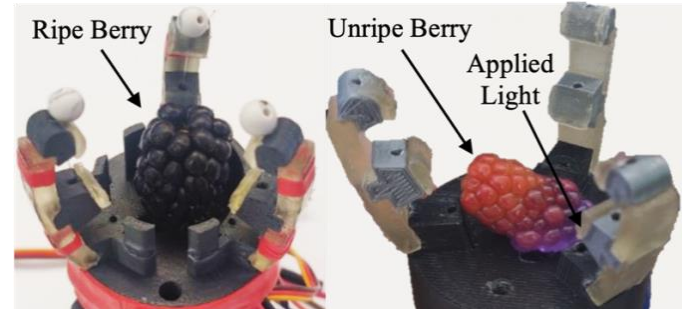

Fig. 5. Left side shows reflectance detection procedure for ripe blackberry and right side shows unripe blackberry being tested. The ripeness sensor composed of LEDs and a photodiode are integrated into the palm of the shown gripper.

### B. Field Tests

Two separate field tests were conducted to evaluate the soft robotic gripper capabilities. The first field test was conducted at a private blackberry farm in Tifton, GA, where the soft robotic gripper underwent blackberry harvesting testing, specifically detachment success rate. This testing was done prior to NIR integration due to the limited blackberry harvesting timeline. However, these blackberries were used in the lab for the experimental methods in Section III-A.

The second field test was conducted at a pick-your-own blackberry farm near Atlanta, GA, and investigated three different harvesting metrics of the gripping and harvesting procedure: (i) retention force required to remove fully ripe and unripe blackberries from their stems, (ii) blackberry ripeness identification through reflectance, and (iii) gripper harvesting efficacy. All three metrics were determined in a sequential manner for each harvested blackberry. The gripper was mounted to a Vernier Go Direct Force and Acceleration Sensor (GDX-FOR, Vernier, USA) on a linear rail (Fig. 6(top)). The force sensor was used to record the retention force of the blackberry with respect to the plant. The harvesting procedure proceeded as follows. First, the gripper was manually located in front of the blackberry so that the blackberry was within the gripper's workspace (Fig 6(middle)). The gripper was actuated, ensuring contact between the sensor at the palm of the gripper and the blackberry (Fig. 6(bottom)). The reflectance data was recorded, confirming the blackberry was ripe. Once the blackberry was confirmed to be ripe, the blackberry was harvested using the gripper. The peak force value recorded during harvesting was selected as the retention force. Following harvesting, the blackberry was placed in a plastic clamshell container for storage and post-harvest marketability evaluation.

Due to the low retention force of the gripper (< 4N), multiple clusters of unripe blackberries still attached to their stems were manually pruned and brought to the lab for benchtop testing two hours after pruning. The retention force of the unripe blackberries was evaluated by tying one end of a string around each blackberry and the other end of the string to the Vernier force sensor. Each blackberry was then harvested from the stem by pulling the force sensor away from the stem until the blackberries were detached. The peak force value recorded during harvesting was selected as the retention force. Additionally, 50 unripe blackberries were handpicked and brought to the lab for reflectance testing.

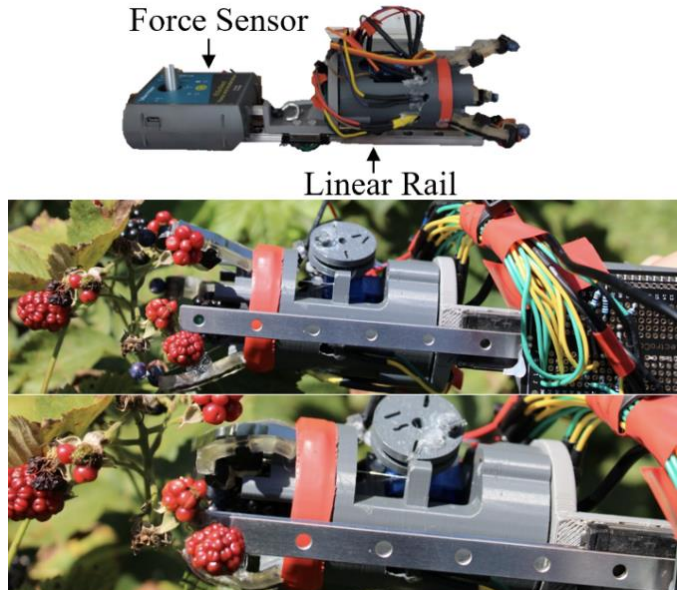

Fig. 6. (Top) shows linear force rail setup with soft robotic gripper. (Middle) shows gripper ready for reflectance detection with blackberry contacting sensor at palm. (Bottom) shows gripper fingers fully enclosing blackberry to conduct harvesting and retention force testing.

### C. UR5 Arm Integration Test

The aim of the UR5 arm implementation test was to provide the proof of concept for automated blackberry harvesting and laid the groundwork for how full system integration would operate with the fabricated soft robotic gripper. Using a 3D printed mount, the soft robotic gripper was rigidly attached to the wrist of the UR5 arm. Three plastic artificial blackberries were set up in front of the soft robotic gripper so that each one was visible in the endoscopic camera frame (Fig. 7). After harvesting a blackberry, the gripper would drop the blackberry into a clamshell container and return to its original position. The process would then be repeated until no more blackberries remained per the protocol in Fig. 4. Note that ripeness sensing was not included in the pipeline due to the use of plastic blackberries.

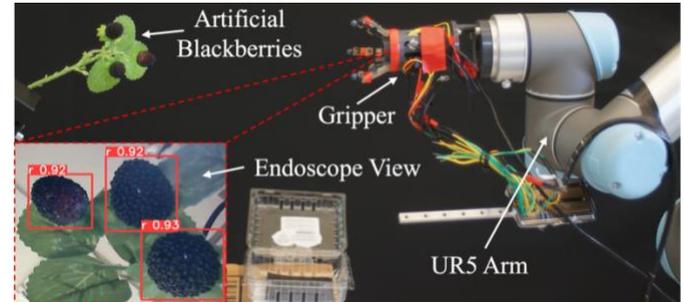

Fig. 7. UR5 arm integration test setup. Gripper must harvest three plastic artificial blackberries with no human interference. The bounding box label values in the endoscope view denote blackberry detection confidence level.

## IV. Results and Discussion

### A. Reflectance Detection

Using 50 ripe blackberries and 25 unripe blackberries collected from the first field test at a private farm in Tifton, GA, the average reflectance for ripe and unripe blackberries was 17.96 and 22.13, respectively. These results showed a distinction between the two different ripeness levels, and proves useful in implementation of the sensor when gripping. With a given threshold in the sensor, classifications can be made to the blackberries before harvesting to determine the course of action. These reflectance values were used to determine ripe and unripe blackberries in the following section.

### B. Field Tests

A total of 139 ripe blackberries were harvested in both field tests. There were 19 failed harvesting attempts (i.e., blackberry remained on the stem). This resulted in a gripper harvesting efficiency of 88%. The force sensor and NIR sensor were used for the 50 blackberries harvested in the second field test. These results can be seen in Table 1. The average retention force for these blackberries was 2.06 N with a standard deviation of 0.92 N. 10 unripe blackberries were pruned given the grower's permission. For the unripe blackberries, the average retention force was 6.08 N with a standard deviation of 1.25 N. Note that the minimum retention force for fully ripe blackberries can be nearly zero. This is due to the ease of detachment in overripened blackberries.

The ripe blackberries tested for reflectance in the field had an average reflectance of 16.78. The 50 handpicked unripe blackberries yielded an average reflectance of 21.70. Figure 8 shows a plot of all 50 ripe and 50 unripe blackberries and their respective reflectance values. This data supports the characterized NIR sensor results in Section IV-A, indicating a useful tool for identifying ripe versus unripe blackberries. Although blackberries of different species may yield different reflectance values, the ripeness sensor may be calibrated as done in these tests before utilization.

TABLE 1: BLACKBERRY RETENTION FORCE RESULTS

| Parameter | Ripe Blackberries | Unripe Blackberries |
|---|---|---|
| Min. $F_r$ [N] | 0.03 | 3.00 |
| Max. $F_r$ [N] | 4.50 | 7.98 |
| Avg. $F_r$ [N] | 2.06 | 6.08 |
| Avg. Length [mm] | 20.30 | 19.58 |
| Avg. Width [mm] | 17.90 | 16.00 |

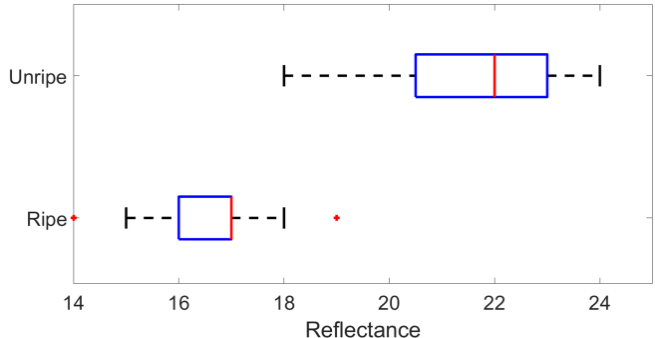

Fig. 8. 870 nm wavelength reflectance results for ripe and unripe blackberries collected at Mabry Farm field testing.

## C. UR5 Arm Integration Test

With a fully implemented artificial blackberry picking pipeline, three artificial blackberries were successfully harvested consecutively in five separate experiments with no human intervention. Each blackberry was successfully harvested without any drops. Fig. 9 shows a 3D plot of the gripper trajectory with respect to the UR5 base coordinate frame during these tests. Each three-blackberry harvesting experiment required five minutes of operation. This is largely due to the speed of the visual servoing adjustments and ethernet communication between laptop and UR5 control box. These times can be significantly reduced by using higher speed gain (user-defined constant that controls magnitude of arm adjustments) and higher error threshold. It should be noted that although this test was only done with three blackberries per experiment, the software developed was capable of harvesting any number of blackberries due to its conditional loop design with a stopping parameter of zero identified blackberries in the camera frame. This test laid the groundwork for fully implementing the soft robotic gripper system as a novel proof of concept test for the legitimate utilization of a tendon-driven soft robotic gripper for automated blackberry harvesting.

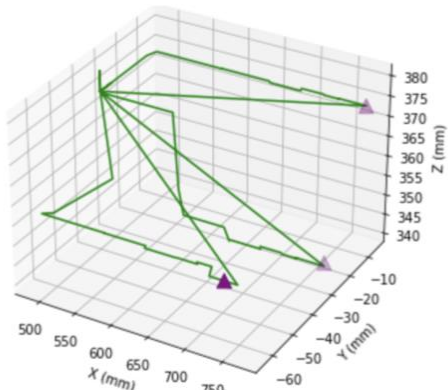

Fig. 9. 3D plot showing gripper position trajectory during visual servoing with respect to UR5 base frame coordinate system.

## V. Conclusions

This paper presented a small form factor and low-cost tendon-driven soft robotic gripper capable of blackberry ripeness sensing and eye-in-hand visual servoing. The soft robotic gripper was field tested with blackberries in Georgia, USA, and had an 88% blackberry detachment success rate. The ripeness sensor utilized a near-infrared wavelength of 870 nm for detecting reflectance from blackberries and successfully determined a clear difference between fully ripe and unripe blackberries during field testing, with fully ripe and unripe blackberries yielding an average reflectance of 16.78 and 21.70, respectively. Finally, the soft robotic gripper was mounted onto a UR5 robot arm and successfully harvested three artificial blackberries utilizing image-based visual servoing under five minutes.

Future work for this project includes: (i) full integration of soft robotic gripper onto a suitable robot arm manipulator and mobile platform for field harvesting tests, (ii) optimizing the gripper for harvesting blackberries found in clusters where they are in close proximity to each other, and (iii) further NIR sensor optimization to improve sensor sensitivity, enabling distinctions between fully ripe and near-ripe blackberries, two ripeness stages that are difficult to differentiate solely with computer vision relying on visible light (the work described in this paper is an initial proof of concept for distinguishing between unripe and ripe blackberries). Additionally, we will begin mechanics modeling of our soft robotic gripper to analyze gripper retention force and finger shape.

## VI. Acknowledgements

The authors would like to acknowledge the contributions of Mabry Farm in Marietta, Georgia and S&A Farm in Tifton, Georgia to this research. Their guidance, insight, and labor growing fresh-market blackberries made this work possible. In addition, the authors thank Huaijin Tu for his assistance in the UR5 IBVS implementation.